\documentclass[pdflatex,sn-basic]{sn-jnl}% Basic Springer Nature Reference Style/Chemistry Reference Style

\usepackage{cleveref}
\usepackage{subfigure}
\usepackage{tikz}
\usepackage{program}
\newcommand\myarrow[1][1]{%
  \raisebox{0.04cm}{\scalebox{0.7}{\begin{tikzpicture}\draw[-stealth] (0,0) -- (#1,0);\end{tikzpicture}}}}

\crefname{section}{§}{§§}
\Crefname{section}{§}{§§}

\usepackage{etoolbox}

\jyear{2021}%

\theoremstyle{thmstyleone}%
\theoremstyle{thmstyletwo}%

\theoremstyle{thmstylethree}%

\def\eg{\emph{e.g.}} 
\def\ie{\emph{i.e.}} 
 
\def\etc{\emph{etc.}}

\def\Ours{{RetroWISE}}

\begin{document}

\title[]{Retrosynthesis prediction enhanced by in-silico reaction data augmentation}
%\title[Article Title]{Machine learning for retrosynthesis: generating reactions from millions of compounds}

\author[1]{\fnm{Xu} \sur{Zhang}}\email{xu.zhang@zju.edu.cn}

\author[2]{\fnm{Yiming} \sur{Mo}}\email{yimingmo@zju.edu.cn}

\author*[1]{\fnm{Wenguan} \sur{Wang}}\email{wenguanwang@zju.edu.cn}
% %\equalcont{These authors contributed equally to this work.}

\author*[1]{\fnm{Yi} \sur{Yang}}\email{yangyics@zju.edu.cn}
% %\equalcont{These authors contributed equally to this work.}

%\author[1]{\fnm{First} \sur{Last}}\email{examples@zju.edu.cn}

\affil[1]{\orgdiv{College of Computer Science and Technology}, \orgname{Zhejiang University}, \orgaddress{\city{Hangzhou}, \postcode{310058}, \state{Zhejiang}, \country{China}}}
\affil[2]{\orgdiv{College of Chemical and Biological Engineering}, \orgname{Zhejiang University}, \orgaddress{\city{Hangzhou}, \postcode{310058}, \state{Zhejiang}, \country{China}}}


\abstract{
Recent advances in machine learning (ML) have expedited retrosynthesis research by assisting chemists to design experiments more efficiently. However, all ML-based methods consume substantial amounts of paired training data (\ie,\!~chemical reaction: \textit{product}-\textit{reactant(s)} pair), which is costly to obtain. Moreover, companies view reaction data as a valuable asset and restrict the accessibility to researchers. These issues prevent the creation of more powerful retrosynthesis models due to their data-driven nature. As a response, we exploit easy-to-access \textit{unpaired} data (\ie, one component of \textit{product}-\textit{reactant(s)} pair) for generating in-silico \textit{paired} data to facilitate model training. Specifically, we present \Ours, a self-boosting framework that employs a base model inferred from \textit{real} paired data to perform \textit{in-silico} reaction generation and augmentation using unpaired data, ultimately leading to a superior model. On three benchmark datasets, \Ours~achieves the best overall performance against state-of-the-art models (\eg, +8.6\% top-1 accuracy on the USPTO-50K test dataset). Moreover, it consistently improves the prediction accuracy of rare transformations. These results show that \Ours~overcomes the training bottleneck by in-silico reactions, thereby paving the way toward more effective ML-based retrosynthesis models.}

\keywords{Retrosynthesis, Machine Learning, In-silico Reaction Data Augmentation, Self-boosting Framework}



\maketitle

\section{Introduction}\label{sec1}

Retrosynthesis, the process of identifying precursors for a target molecule, is essential for material design and drug discovery$_{\!}$~\citep{blakemore2018organic}. However, the huge search space for possible chemical transformations and enormous time required even for 
experts make this challenging.
Thus, efficient computer-assisted synthesis $_{\!}$~\citep{corey1969computer,corey1985computer,coley2017computer} has been explored for long periods. Thanks to recent advances in artificial intelligence, machine learning (ML)-based methods$_{\!}$~\citep{segler2018planning,mikulak2020computational,schwaller2021mapping,toniato2021unassisted,yu2023machine,born2023regression} have emerged to assist chemists to design experiments and gain insights that might not be solely achievable through traditional methods, bringing retrosynthesis research to a new pivotal moment. 

The ML-based methods for single-step retrosynthesis can be roughly categorized into three groups: \textbf{Template-based methods} predict reactants using reaction templates that encode core reactive rules. LHASA$_{\!}$~\citep{corey1985computer}, the first retrosynthesis program, utilizes manual-encoding templates to predict retrosynthetic routes. To scale to exponentially growing knowledge$_{\!}$~\citep{segler2018planning}, data-driven methods$_{\!}$~\citep{segler2017neural,coley2017computer,dai2019retrosynthesis,baylon2019enhancing,chen2021deep} extract a large number of reaction templates from data and formulate retrosynthesis as a template retrieval/classification task. 
\textbf{Semi-template methods}$_{\!}$~\citep{shi2020graph,yan2020retroxpert,somnath2021learning,wang2021retroprime} decompose retrosynthesis into two stages: they typically (1) identify the reactive sites to convert the product into synthons and (2) complete the synthons into reactant(s), which utilize ``reaction centers'' in templates to supervise the training procedure$_{\!}$~\citep{sun2021towards}. 
\textbf{Template-free methods} view single-step retrosynthesis prediction as a machine translation task, where deep generative models directly translate the given product into reactant(s). These methods use either SMILES$_{\!}$~\citep{weininger1988smiles} or molecular graph as data representations, leading to sequence-based methods$_{\!}$~\citep{liu2017retrosynthetic,tetko2020state,lin2020automatic,kim2021valid,wan2022retroformer,zhong2022root} and graph-based methods$_{\!}$~\citep{seo2021gta,tu2022permutation,zhong2023retrosynthesis}, respectively.

Despite appealing results, existing ML-based methods have an insatiable appetite for paired training data (\ie, chemical reaction: product-reactant(s) pair), which is costly to obtain since chemistry experiments are typically not designed to build reaction databases but to meet the specific research need$_{\!}$~\citep{rodrigues2019good}. Moreover, chemical reaction collection is time-consuming and requires domain expertise, making it a valuable asset to companies. As a result, proprietary databases (\eg,~Reaxys$_{\!}$~\citep{lawson2014making} collected from scientific literatures and organic chemistry/life science patents) have limited accessibility, which cannot be viewed and acquired directly. In contrast, public datasets such as USPTO$_{\!}$~\citep{lowe2012extraction,Lowe2017} extracted from US patents have finite paired data (roughly $3.7$M reactions with duplicates). These issues remain key obstacles to impede progress toward more effective retrosynthesis models due to their data-driven nature.
In response to such issues, data augmentation with newly generated samples has been a recent success in various fields, such as medical research$_{\!}$~\citep{marouf2020realistic,gao2023synthetic}, biological research$_{\!}$~\citep{castro2022transformer,baker2023silico}, and robotic research$_{\!}$~\citep{yang2022automatic}, as it provides an inexpensive augmentation without increasing the demand for costly data collection and raising privacy concerns. However, the development of in-silico reaction generation and augmentation for single-step retrosynthesis prediction has yet to be explored. 

Here, we present a framework called \Ours~that uses a base model inferred from real paired data to generate in-silico paired data from unpaired data (\ie, one component of the product-reactant(s) pair), which can be more easily collected in public databases or via web scraping, to develop a more effective ML model. Specifically, \Ours~uses real paired data to train the base model, and then generates abundant in-silico reactions from easy-to-access unpaired data using the base model. Finally, \Ours~augments real paired data with the generated reactions to train a more effective retrosynthesis model. In this way, our training ends up in a self-boosting manner: In-silico reactions generated from the base model in turn push the model to evolve. 
We conduct experiments on three widely used benchmark datasets of single-step retrosynthesis prediction. The experimental results provide encouraging evidence that \Ours~achieves the best overall performance against state-of-the-art models (\eg, 8.6\% improvement of top-1 accuracy on the USPTO-50K$_{\!}$~\citep{schneider2016s} test dataset). Moreover, we show that \Ours~consistently promotes the prediction accuracy on rare transformations, which are typically of particular interest to chemists for novel synthetic routes design. In summary, \Ours~provides a  feasible and cost-effective way of in-silico reaction generation and augmentation based on self-boosting procedure to advance the ML-based retrosynthesis research.

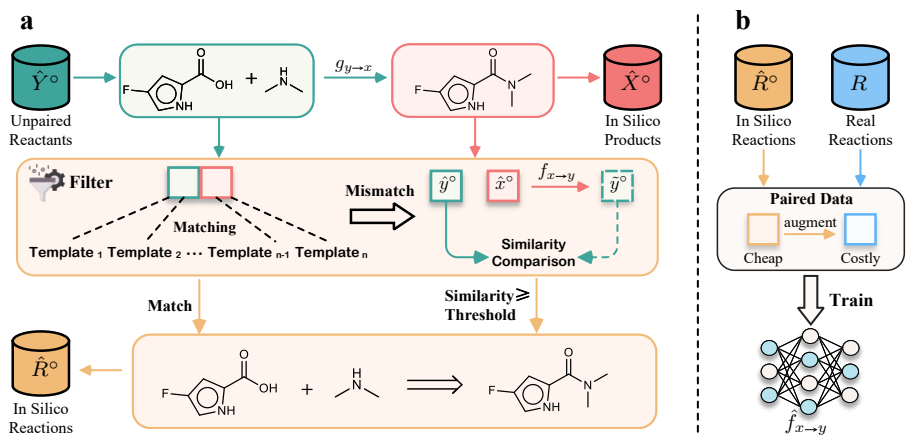
\begin{figure*}[t]%
    \centering
    \includegraphics[width=\textwidth]{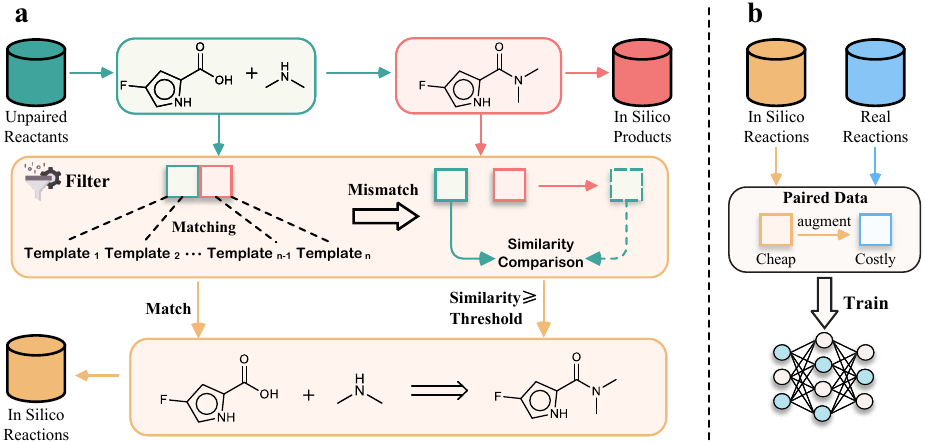}
    \put(-329, 130.5){\fontsize{8pt}{1em}\selectfont $\hat{Y}^\circ$}
    \put(-108, 130.5){\fontsize{8pt}{1em}\selectfont $\hat{X}^\circ$}
    \put(-329, 22){\fontsize{8pt}{1em}\selectfont $\hat{R}^\circ$}
    \put(-215.0, 139){\scalebox{0.7}{$g_{y\myarrow[0.3]x}$}}
    \put(-139, 98){\scalebox{0.7}{$f_{x\myarrow[0.3]y}$}}
    \put(-155, 93){\scalebox{0.7}{$\hat{x}^\circ$}}
    \put(-176, 93){\scalebox{0.7}{$\hat{y}^\circ$}}
    \put(-112, 93){\scalebox{0.7}{$\tilde{y}^\circ$}}
    \put(-22, 130.5){\fontsize{8pt}{1em}\selectfont $R$}
    \put(-59, 130.5){\fontsize{8pt}{1em}\selectfont $\hat{R}^\circ$}
    \put(-46.5, 2){\scalebox{0.8}{$\hat{f}_{x\myarrow[0.3]y}$}}
    \caption{\textbf{Overview of the \Ours~framework. a}, Given the unpaired reactants $\hat{Y}^\circ$ as an example, the base forward synthesis model $g_{y\myarrow[0.3]x}$ trained on real paired data is used to generate in-silico products $\hat{X}^\circ$. Then, a filter process consisting of template matching and molecular similarity comparison selects high-quality in-silico reactions $\hat{R}^\circ$. \textbf{b}, These cheap in-silico reactions are used to augment costly real reactions as paired training data to train a more effective retrosynthesis model $\hat{f}_{x\myarrow[0.3]y}$. In this way, the whole framework is self-boosted.}
    \label{fig:overview}
\end{figure*}

\section{Results}\label{sec:result}

\noindent\textbf{\Ours~framework.} In the presence of real paired data, which consists of product-reactant(s) pairs, and unpaired data (products or reactants), the main idea behind \Ours~is in a self-boosting manner: employing a base model to generate in-silico reactions from unpaired data, which in turn augment real paired data to facilitate model training. Specifically, \Ours~uses real paired data $\mathit{R}\!=\!\{(x_{n},y_{n})\}_{n}$, where $x_{n}$ represents the product and $y_{n}$ denotes the corresponding reactant(s), to train a base forward synthesis model $g_{y\myarrow[0.3]x}$ and a base retrosynthesis model $f_{x\myarrow[0.3]y}$ as the preparation. Then, as illustrated in Fig.\!~\ref{fig:overview}a, \Ours~generates in-silico reactions in one of two ways: (1) using the base forward synthesis model $g_{y\myarrow[0.3]x}$ to produce in-silico products $\hat{\mathit{X}}^{\circ}\!=\!\{\hat{x}^\circ_{m}\}_{m}$ from unpaired reactants $\hat{\mathit{Y}}^{\circ}\!=\!\{\hat{y}^{\circ}_{m}\}_{m}$; (2) using the base retrosynthesis model $f_{x\myarrow[0.3]y}$ to generate in-silico reactants $\hat{\mathit{Y}}^{\star}\!=\!\{\hat{y}^{\star}_{l}\}_{l}$ from unpaired products $\hat{\mathit{X}}^{\star}\!=\!\{\hat{x}^{\star}_{l}\}_{l}$. The unpaired data (\eg, unpaired reactant(s)) and in-silico data (\eg, in-silico product) make up each generated reaction. Moreover, to enhance the quality of in-silico reactions, \Ours~incorporates a filter process with chemical awareness, which consists of a template matching step and a molecular similarity comparison step: (1) preserving generated reactions matching any selected template; (2) reconstructing pseudo unpaired data (\eg, $\tilde{y}^{\circ}$) from in-silico data (\eg, $\hat{x}^\circ$) of the mismatched reactions from the previous step, comparing the molecular similarity to the original unpaired data (\eg, $\hat{y}^{\circ}$), and retaining in-silico reactions with the molecular similarity above a specific threshold. For brevity, the preserved in-silico reactions $\mathit{\hat{R}}^{\circ}\!=\!\{(\hat{x}^{\circ}_{m},\hat{y}^{\circ}_{m})\}_{m}$ and $\mathit{\hat{R}}^{\star}\!=\!\{(\hat{x}^{\star}_{l},\hat{y}^{\star}_{l})\}_{l}$ are denoted as $\mathit{\hat{R}}$. Finally, as illustrated in Fig.\!~\ref{fig:overview}b, \Ours~uses cheap in-silico paired data $\mathit{\hat{R}}$ to augment costly real paired data $\mathit{R}$ to train a more powerful retrosynthesis model $\hat{f}_{x\myarrow[0.3]y}$.

\noindent\textbf{Improvements using in-silico reactions.} Learning from sufficient paired training data is a key factor in the success of ML-based retrosynthesis methods. Thereby, we investigate how \Ours~improves the retrosynthesis prediction performance by augmenting paired training data with in-silico reactions. First, \Ours~generates in-silico reactions from unpaired reactants in USPTO applications$_{\!}$~\citep{Lowe2017}. Specifically, the raw reactions are preprocessed as in$_{\!}$~\cite{dai2019retrosynthesis} to obtain approximately $1$M unique reactants. Then, \Ours~utilizes the base forward synthesis model $g_{y\rightarrow x}$ to produce the corresponding in-silico products $\hat{\mathit{X}}^{\circ}$ from the unpaired reactants $\hat{\mathit{Y}}^{\circ}$, and forms them as in-silico reactions $\hat{\mathit{R}}^{\circ}$. \Ours~trained with extra generated reactions from USPTO applications is referred to as \Ours-\textit{U}. This generation and training procedure is particularly useful when having plentiful reactants without knowing the outcomes in advance. 
Second, \Ours~obtains in-silico reactions from unpaired products by randomly sampling $4$M molecules from the PubChem database$_{\!}$~\citep{kim2019pubchem} or $20$M from the ZINC database$_{\!}$~\citep{irwin2020zinc20}. \Ours~utilizes the base retrosynthesis model $f_{x\rightarrow y}$ to produce in-silico reactants $\hat{\mathit{Y}}^{\star}$ from the unpaired products $\hat{\mathit{X}}^{\star}$, and form them as in-silico reactions $\hat{\mathit{R}}^{\star}$. For better differentiation, we denote \Ours~trained with extra generated reactions from PubChem and ZINC as \Ours-\textit{P} and \Ours-\textit{Z}, respectively. This pipeline is also feasible as numerous molecules are publicly accessible in large databases.% in public databases. 

\begin{table}[t]
    \caption{Performance (\%) of various models trained with in-silico reactions from different unpaired data sources on the USPTO-50K test set and the USPTO-MIT test set.}
    \label{tab:improvement}
    \setlength\tabcolsep{12pt}
      \resizebox{\linewidth}{!}{
    \begin{tabular}{@{}lcccc}
    \toprule
    \multirow{2}{*}{Model} &\multirow{2}{*}{Extra paired data} & \multicolumn{2}{c}{USPTO-50K} & USPTO-MIT \\ \cmidrule(l){3-5} 
           && Top-1    & Top-1 $\left\langle MaxFrag\right\rangle$    & Top-1 \\ \midrule
    Baseline& None   & 56.3   &61.0   &60.3   \\ 
    \Ours-\textit{P}&  4M    & 60.0  & 64.1            &61.6    \\
    \Ours-\textit{Z}&  20M    & 60.1  & 64.7            &61.9    \\
    \Ours-\textit{U}&  320K  & \textbf{63.8}  & \textbf{68.5}         &\textbf{64.6}   \\

    \bottomrule
    \end{tabular} }
\end{table}

As shown in Table~\ref{tab:improvement}, we evaluate our models (\ie, \Ours-\textit{U}, \Ours-\textit{P}, and \Ours-\textit{Z}) on two benchmark datasets: USPTO-50K$_{\!}$~\citep{schneider2016s} and USPTO-MIT$_{\!}$~\citep{jin2017predicting}. The baseline is trained only with real paired data, while \Ours~is trained with the same real paired data, as well as in-silico reactions as auxiliary paired training data. The evaluation metrics are top-1 exact accuracy and top-1 MaxFrag accuracy$_{\!}$~\citep{tetko2020state}. On USPTO-50K, \Ours-\textit{U} achieves the highest exact match accuracy at 63.8\% and the highest MaxFrag accuracy at 68.5\%. \Ours-\textit{P} and \Ours-\textit{Z} also have clear advantages over the baseline, yielding enhancements of 3.5\% and 3.8\% on top-1 exact accuracy. Moreover, \Ours~achieves significant improvements on the larger USPTO-MIT dataset, \eg, \Ours-\textit{U}, \Ours-\textit{P}, and \Ours-\textit{Z} exceed the baseline by 4.3\%, 1.6\%, and 1.3\%, respectively. We attribute the superior performance of \Ours-\textit{U} to two factors: (1) the base forward synthesis model $g_{y\myarrow[0.3]x}$ used to generate in-silico data is much more accurate than the base retrosynthesis model  $f_{x\myarrow[0.3]y}$, producing higher-quality reactions; and (2) unpaired reactants in \Ours-\textit{U} are more likely to generate chemically plausible reactions than randomly sampled unpaired products from PubChem and ZINC. These results show that incorporating in-silico reactions indeed facilitate model learning process and that \Ours~provides better results by using in-silico reactions generated from unpaired reactants.

\begin{table}[t]
        \centering
        \caption{Top-\textit{k} single-step retrosynthesis accuracy (\%) on the USPTO-50K test set.}
        \label{tab:topk-50K}
        \resizebox{\textwidth}{!}{%
        \begin{tabular}{@{}llcccccc@{}}
        \toprule
        Category       & Model                     & \textit{k} = 1 & 3    & 5    & 10   & 20   & 50   \\ \midrule
        Template-based & Retrosim$_{\!}$~\citep{coley2017computer}   & 37.3  & 54.7 & 63.3 & 74.1 & 82.0 & 85.3 \\
                       & Neuralsym$_{\!}$~\citep{segler2017neural}   & 44.4  & 56.3 & 72.4 & 78.9 & 82.2 & 83.1 \\
                       & GLN$_{\!}$~\citep{dai2019retrosynthesis}    & 52.5  & 69.0 & 75.6 & 83.7 & 89.0 & 92.4 \\
                       & LocalRetro$_{\!}$~\citep{chen2021deep}      & 53.4  & 77.5 & 85.9 & 92.4 & -    & 97.7 \\ \midrule
        Semi-template  & G2Gs$_{\!}$~\citep{shi2020graph}            & 48.9  & 67.6 & 72.5 & 75.5 & -    & -    \\
                       & GraphRetro$_{\!}$~\citep{somnath2021learning} & 53.7  & 68.3 & 72.2 & 75.5 & -    & -    \\
                       & RetroXpert\footnotemark[1]$_{\!}$~\citep{yan2020retroxpert} & 50.4  & 61.1 & 62.3 & 63.4 & 63.9 & 64.0 \\
                       & RetroPrime$_{\!}$~\citep{wang2021retroprime} & 51.4  & 70.8 & 74.0 & 76.1 & -    & -    \\ \midrule
        Template-free  & Liu's~Seq2seq$_{\!}$~\citep{liu2017retrosynthetic} & 37.4  & 52.4 & 57.0 & 61.7 & 65.9 & 70.7 \\
                       & GTA$_{\!}$~\citep{seo2021gta}               & 51.1  & 67.6 & 74.8 & 81.6 & -    & -    \\
                       & Dual-TF$_{\!}$~\citep{sun2021towards}       & 53.3  & 69.7 & 73.0 & 75.0 & -    & -    \\
                       & MEGAN$_{\!}$~\citep{sacha2021molecule}      & 48.1  & 70.7 & 78.4 & 86.1 & 90.3 & 93.2 \\
                       & Tied transformer$_{\!}$~\citep{kim2021valid} & 47.1  & 67.2 & 73.5 & 78.5 & -    & -    \\
                       & AT$_{\!}$~\citep{tetko2020state}            & 53.5  & -    & 81.0 & 85.7 & -    & -    \\
                       & Graph2Edits$_{\!}$~\citep{zhong2023retrosynthesis}    & 55.1  & 77.3    & 83.4 & 89.4 & -    & 92.7    \\
                       & R-SMILES$_{\!}$~\citep{zhong2022root}       & 56.3  & 79.2 & 86.2 & 91.0 & 93.1 & 94.6 \\
                       & \Ours~(This work) &\textbf{64.9}	&\textbf{83.5}	&\textbf{88.4}	&\textbf{92.7}	&\textbf{95.1}	& \textbf{96.9} \\
                       \cmidrule(l){2-8}
                       %& \textbf{MaxFrag match} &   &  &  &  &  &  \\
        Template-free  & MEGAN           & 54.2  & 75.7 & 83.1 & 89.2 & 92.7 & 95.1 \\
        $\left\langle MaxFrag\right\rangle$
                       & Tied transformer   &  51.8  & 72.5 & 78.2 &  82.4 & - & - \\
                       & AT              & 58.5  & -    & 95.4 & 90.0 & -    & -    \\
                       & Graph2Edits    & 59.2  & 80.1    & 86.1 & 91.3 & -    & 93.1    \\
                       & R-SMILES         & 61.0  & 82.5 & 88.5 & 92.8 & 94.6 & 95.7 \\
                       & \Ours~(This work)  &\textbf{69.1}	&\textbf{86.5}	&\textbf{90.4}	&\textbf{93.6}	&\textbf{95.5}	&\textbf{97.0} \\
        \botrule
        \end{tabular}}
        \footnotetext[1]{\href{https://github.com/uta-smile/RetroXpert.}{RetroXpert} results are updated by the official implementation}
    \end{table}
    
    \begin{table}[t]
        \centering
        \caption{Top-\textit{k} single-step retrosynthesis accuracy (\%) on the USPTO-MIT test set.}
        \label{tab:topk-mit}
        \resizebox{\textwidth}{!}{%
        \begin{tabular}{@{}llcccccc@{}}
        \toprule
        Category       & Model                     & \textit{k} = 1 & 3    & 5    & 10   & 20   & 50   \\ \midrule
        Template-based & Neuralsym$_{\!}$~\citep{segler2017neural}   & 47.8  & 67.6 & 74.1 & 80.2 & - & - \\
                       & LocalRetro$_{\!}$~\citep{chen2021deep}      & 54.1  & 73.7 & 79.4 & 84.4 & -    & 90.4 \\ \midrule
        Template-free  & Liu's~Seq2seq$_{\!}$~\citep{liu2017retrosynthetic} & 46.9  & 61.6 & 66.3 & 70.8 & - & - \\
                       & AutoSynRoute$_{\!}$~\citep{lin2020automatic} & 54.1  & 71.8 & 76.9 & 81.8 & -    & -    \\
                       & RetroTRAE$_{\!}$~\citep{ucak2022retrosynthetic} & 58.3  & - & - & - & -    & -    \\
                       & R-SMILES$_{\!}$~\citep{zhong2022root}  & 60.3  & 78.2 & 83.2 & 87.3 & 89.7 & 91.6 \\
                       & \Ours~(This work)  &\textbf{64.6}	&\textbf{82.3}	&\textbf{86.7}	&\textbf{90.3}	&\textbf{92.4}	&\textbf{94.0}   \\
        \botrule
        \end{tabular}    }
    \end{table}
    
    \begin{table}[h]
        \centering
        \caption{Top-\textit{k} single-step retrosynthesis accuracy (\%) on the USPTO-Full test set.}
        \label{tab:topk-full}
        \resizebox{\textwidth}{!}{%
        \begin{tabular}{@{}llcccccc@{}}
        \toprule
        Category       & Model                     & \textit{k} = 1 & 3    & 5    & 10   & 20   & 50   \\ \midrule
        Template-based & Retrosim$_{\!}$~\citep{coley2017computer}   & 32.8  & - & - & 56.1 & - & - \\
                       & Neuralsym$_{\!}$~\citep{segler2017neural}   & 35.8  & - & - & 60.8 & - & - \\
                       & GLN$_{\!}$~\citep{dai2019retrosynthesis}    & 39.3  & - & - & 63.7 & - & - \\
                       & LocalRetro$_{\!}$~\citep{chen2021deep}      & 39.1  & 53.3 & 58.4 & 63.7 & 67.5   & 70.7 \\ \midrule
        Semi-Template  & RetroPrime$_{\!}$~\citep{wang2021retroprime} & 44.1  & 59.1 & 62.8 & 68.5 & -    & -    \\ \midrule
        Template-free  & MEGAN$_{\!}$~\citep{sacha2021molecule}      & 33.6  & - & - & 63.9 & - & 74.1 \\
                       & GTA$_{\!}$~\citep{seo2021gta} & 46.6  & - & - & 70.4 & -    & -    \\
                       & AT$_{\!}$~\citep{tetko2020state}            & 46.2  & -    & - & 73.3 & -    & -    \\
                       & R-SMILES$_{\!}$~\citep{zhong2022root}       & 48.9  & 66.6 & 72.0 & 76.4 & 80.4 & 83.1 \\
                       %& Baseline                &51.4	&67.5	 &72.0	  &76.4	&79.4	&82.2   \\
                       & \Ours~(This work)                     &\textbf{52.3}	&\textbf{68.7}	&\textbf{73.5}	&\textbf{77.9}	&\textbf{80.9}	&\textbf{83.6}  \\
        \botrule
        \end{tabular}    }
    \end{table}
 
\noindent\textbf{Comparison with existing ML-based methods.} Here, we compare \Ours~with other ML-based methods using the most popular retrosynthesis benchmark datasets: USPTO-50K$_{\!}$~\citep{schneider2016s}, USPTO-MIT$_{\!}$~\citep{jin2017predicting}, and USPTO-Full$_{\!}$~\citep{dai2019retrosynthesis}. The top-\textit{k} exact match accuracy and the top-\textit{k} MaxFrag accuracy$_{\!}$~\citep{tetko2020state} are adopted as the evaluation metrics. 
The performance of our \Ours~are summarized in Table~\ref{tab:topk-50K},~\ref{tab:topk-mit}, and~\ref{tab:topk-full}, from which we could derive three critical observations:

\begin{enumerate}
    \item The proposed \Ours~framework outperforms existing state-of-the-art methods$_{\!}$~\citep{chen2021deep,wang2021retroprime,zhong2022root} on all the three datasets, \eg, \Ours~surpasses R-SMILES by 8.6\%, 4.3\%, and 3.4\% top-1 exact accuracy on USPTO-50K, USPTO-MIT, and USPTO-Full, respectively. Our method constantly achieves higher accuracy rates across all top-\textit{k} accuracies, which attests to its effectiveness in tackling the complex single-step retrosynthesis prediction task.
    \item \Ours~is superior to the other methods especially in the low-resource setting with limited paired data. Notably, our method achieves substantial improvements over R-SMILES on USPTO-50K, with an absolute increase of 8.6\%, 4.3\%, and 2.2\% in top-1, top-3, and top-5 accuracies, respectively. These results further confirm the effectiveness of \Ours~by in-silico reaction augmentation under limited resource circumstance.
    \item \Ours~also delivers best results in top-\textit{k} MaxFrag accuracy across the three datasets. The MaxFrag accuracy proposed by$_{\!}$~\cite{tetko2020state} reflects the accuracy to predict the minimal part of reactant(s) for designing a retrosynthetic route, emphasizing multiple possible ways to synthesize the compounds$_{\!}$~\citep{dubrovskiy2018comprehensive}. The highest top-\textit{k} MaxFrag accuracy (\eg, +8.1\% top-1 MaxFrag accuracy on USPTO-50K) underscores the prediction diversity as well as prediction accuracy of \Ours.
\end{enumerate}

\begin{figure*}[t]%
    \centering
    \includegraphics[width=\textwidth]{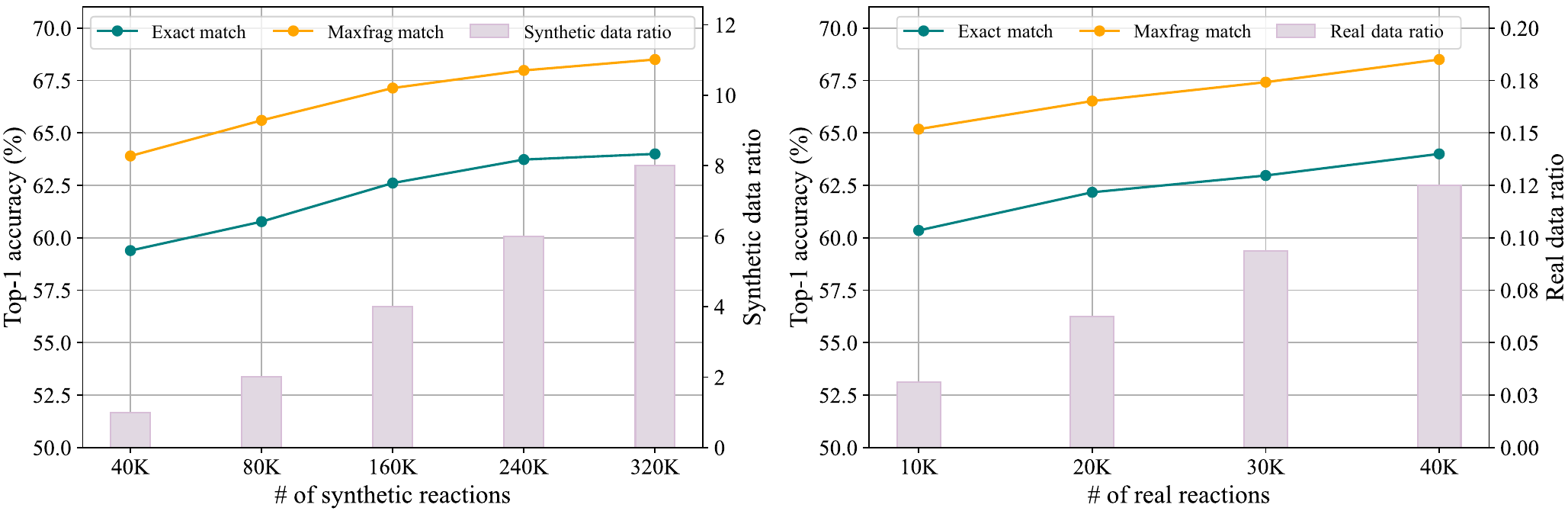}
    \put(-319, 112){\fontsize{6.5pt}{1em}\selectfont \textbf{a}}
    \put(-149, 112){\fontsize{6.5pt}{1em}\selectfont \textbf{b}}
    \caption{\textbf{Impact of data quantity. a,} impact of in-silico data quantity and \textbf{b,} impact of real data quantity. Training with more in-silico and real data both improves the performance. The in-silico data ratio is measured as the number of in-silico reactions divided by the number of real data, and vice versa for the real data ratio.}\label{fig:data_ratio}
\end{figure*}

\noindent\textbf{Impact of data quantity.} The quantity of the paired training data really matters. Next, we will evaluate how \Ours's~performance scales \textit{w.r.t.} amount of data used. We first investigate how the amount of in-silico reactions $\mathit{\hat{R}}$ affects prediction accuracy. A series of experiments are conducted on USPTO-50K$_{\!}$~\citep{schneider2016s} where the number of in-silico reactions $\mathit{\hat{R}}$ is gradually increased. Fig.\!~\ref{fig:data_ratio}a demonstrates that more in-silico reactions lead to higher accuracy for each \textit{k}-value. For instance, the top-1 accuracy increases from around 59.3\% to 63.8\% and top-1 MaxFrag accuracy rises from 63.9\% to 68.5\% when more in-silico reactions are used. It could also be observed that the prediction accuracy continues to grow as $\mathit{\hat{R}}$ increases. This suggests that increasing the amount of in-silico reactions indeed benefits the model training. In turn, we examine the effect of the amount of real paired data on prediction performance. We fix the size of in-silico reactions $\mathit{\hat{R}}$ to $320$K and alter the size of real paired data $\mathit{R}$ in the range of \{$10$K, $20$K, $30$K, $40$K\}. As shown in Fig.\!~\ref{fig:data_ratio}b, we observe that a larger data size of real reactions also leads to higher accuracy, \eg, the top-1 accuracy rises from 60.3\% to 63.8\% as the size of $\mathit{R}$ increases from $10$K to $40$K. These results highlight the importance of increasing the data quantity for training a powerful retrosynthesis model. 

\begin{table}[t]
    \caption{Filter process raises the quality of in-silico reactions for better performances.} 
    \label{tab:filter}
    \setlength\tabcolsep{13pt}
      \resizebox{\linewidth}{!}{
    \begin{tabular}{@{}lcccccc}
    \toprule

    Method & \textit{k} = 1 & 3    & 5    & 10   & 20   & 50   \\ \midrule
    Baseline & 56.3  & 79.2 & 86.2 & 91.0 & 93.1 & 94.6 \\
    \Ours~(\textit{w/o} filtering)    &63.8	&83.0	&87.6	&91.7	&94.1	&95.1 \\
    \Ours~(\textit{w} filtering)     &\textbf{64.9}	&\textbf{83.5}	&\textbf{88.4}	&\textbf{92.7}	&\textbf{95.1}	& \textbf{96.9} \\

    \bottomrule
    \end{tabular} }
\end{table}

\noindent\textbf{Impact of data quality.} Erroneous or low-quality in-silico reactions might result in error accumulation during model training. To address this issue, \Ours~is equipped with a filter process that leverages template matching and molecular similarity comparison to enhance the quality of in-silico reactions. Initially, the filter employs RDKit$_{\!}$~\citep{landrum2013rdkit} to eliminate in-silico reactions that contain wrong reactants or products SMILES. Subsequently, the template matching step selects chemical templates extracted with RDChiral$_{\!}$~\citep{coley2019rdchiral} that appear more than $5$ times ($14.5\%$ of $301,257$ templates in USPTO) as a template library, and then preserves in-silico reactions that match any selected template in this library. This procedure ensures the chemical plausibility of in-silico reactions. Next, the molecular similarity comparison step (1) reconstructs pseudo unpaired data from in-silico data of the mismatched reactions from the last step and (2) uses RDKit to calculate the molecular similarity between the pseudo unpaired data and the original unpaired data. Specifically, as illustrated in Fig.\!~\ref{fig:overview}(a), we feed the in-silico data (\eg, in-silico product $\hat{x}^{\circ}$) into the base model (\eg, $f_{x\myarrow[0.3]y}$) to generate the pseudo unpaired data (\eg, pseudo reactant(s) $\tilde{y}^{\circ}$) for comparing the molecular similarity to original unpaired data (\eg, unpaired reactant(s) $\hat{y}^{\circ}$). Reactions with similarity above a specific threshold are also preserved. This procedure considers the diversity of in-silico reactions while ensuring data validity. As shown in Table~\ref{tab:filter}, the filter process further improves the top-\textit{k} prediction performance of \Ours~(\eg, +1.1\% top-1 accuracy and +1.8\% top-50 accuracy over the baseline) with less in-silico reactions (89\% of the original in-silico paired data). The results verify the rationality of leveraging the filter process for improving data quality such that our \Ours~could benefit from the correct and chemically sound in-silico reactions.

\begin{figure}[t]
    \centering
    \includegraphics[width=\textwidth]{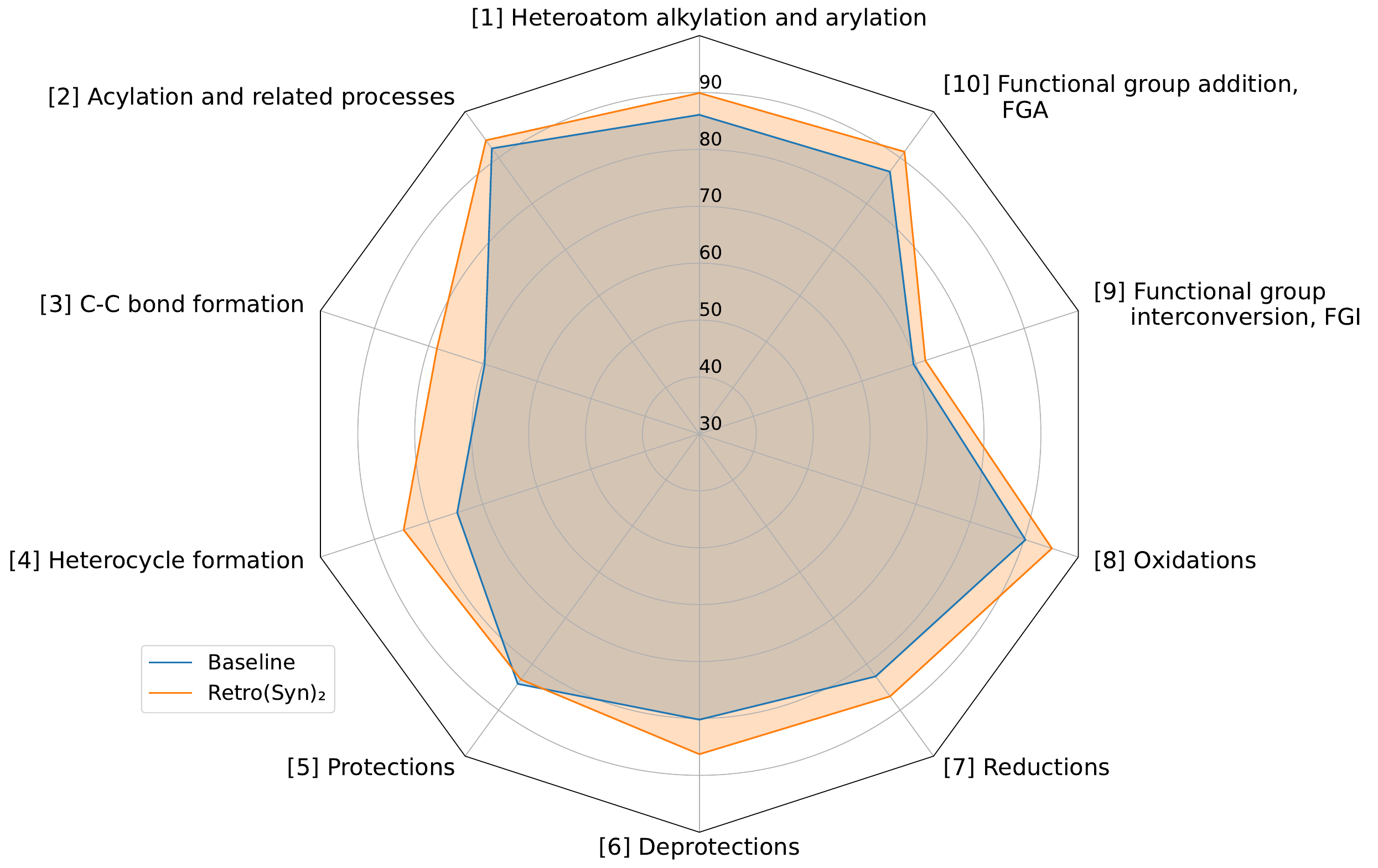}
    \caption{\textbf{Top-5 accuracy of different types of predictions.} \Ours~achieves excellent results over the baseline on almost every reaction type.}\label{fig:ten_class}
\end{figure}

\noindent\textbf{Performance on different reaction types.} Reaction types are crucial to chemists as they usually use them to navigate large databases of reactions and retrieve similar members of the same class to analyze and infer optimal reaction conditions. They also use reaction types as an efficient way to communicate what a chemical reaction does and how it works in terms of atomic rearrangements. Thereby, it is necessary to analyze the performance of different reaction types using the USPTO-50K dataset$_{\!}$~\citep{schneider2016s}, which assigns one of ten reaction classes to each reaction. These classes cover the most common reactions in organic synthesis, such as protections/deprotections, C-C bond formation, and heterocycle formation. Note that, \Ours~does not use reaction types for training since they are often unavailable in real-world scenarios. However, as shown in Fig.\!~\ref{fig:ten_class}, \Ours~outperforms the baseline on almost every reaction type by a large margin. Also, we find that our \Ours~significantly enhances heterocycle formation and C-C bond formation prediction among the ten reaction types (\eg, $9.8\%$ improvement on heterocycle formation class), while protections is the most challenging to predict. We infer that the reasons are (1) heterocycle formation and C-C bond formation have more diverse possibilities for choosing reactants and reactions than other reaction types$_{\!}$~\citep{tetko2020state}; (2) in-silico reactions of protections appear less frequently, resulting in a slight imbalance during model learning. 

\begin{figure*}[t]
    \centering
    \includegraphics[width=\textwidth]{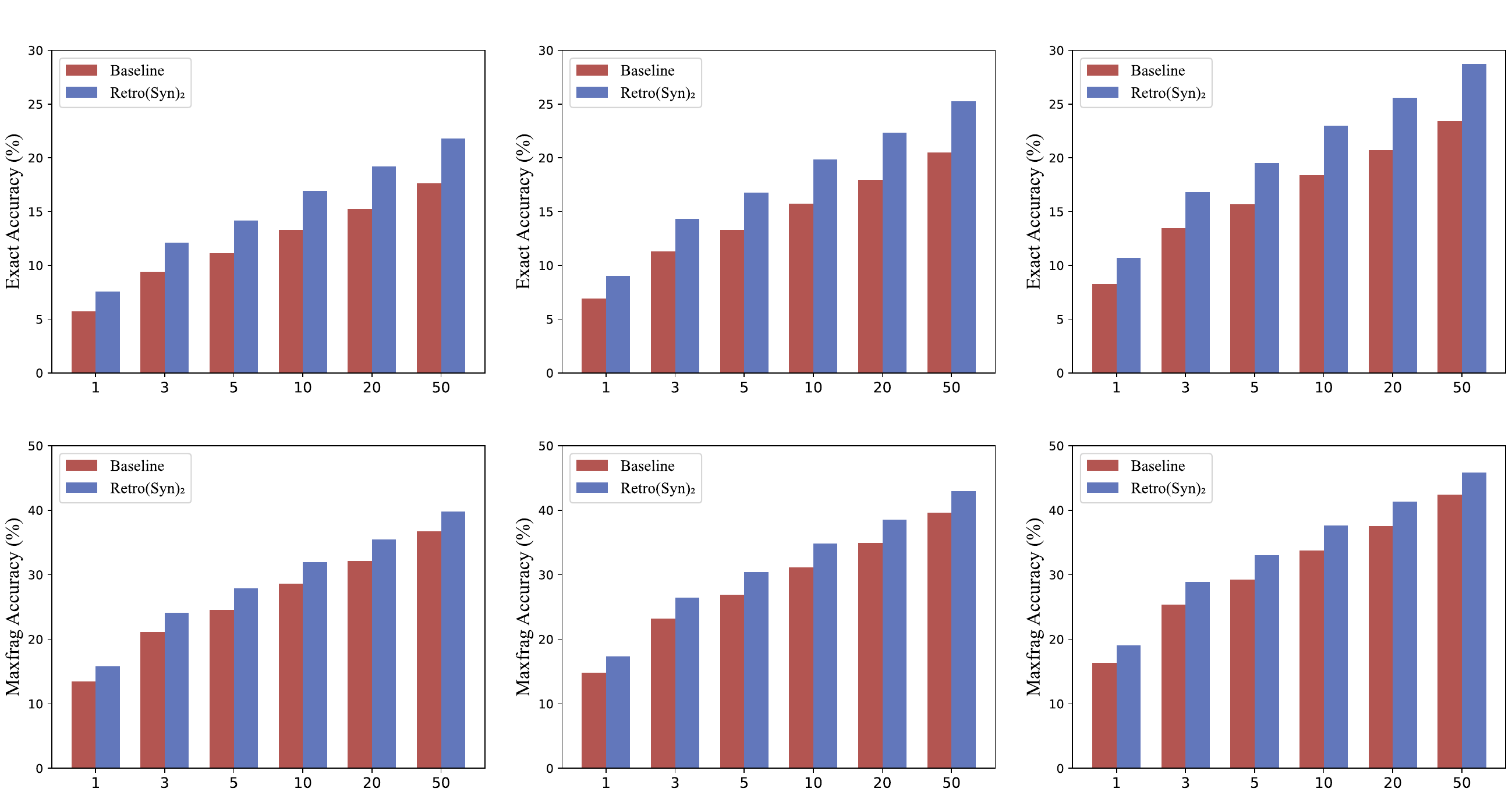}
    \put(-340, 170){\fontsize{6.5pt}{1em}\selectfont \textbf{a}}
    \put(-340, 85){\fontsize{6.5pt}{1em}\selectfont \textbf{b}}
    \put(-292, 168){\fontsize{6.5pt}{1em}\selectfont Rare-2}
    \put(-177, 168){\fontsize{6.5pt}{1em}\selectfont Rare-5}
    \put(-62, 168){\fontsize{6.5pt}{1em}\selectfont Rare-10}
    \caption{\textbf{Performance on rare transformations.} \textbf{a}, Top-\textit{k} exact match accuracy. \textbf{b}, Top-\textit{k} MaxFrag match accuracy. \Ours~achieves consistent improvements on three testing benchmarks of rare transformations.}\label{fig:rare_case}
\end{figure*}

\begin{figure*}[ht]
    \centering
    \includegraphics[width=\textwidth]{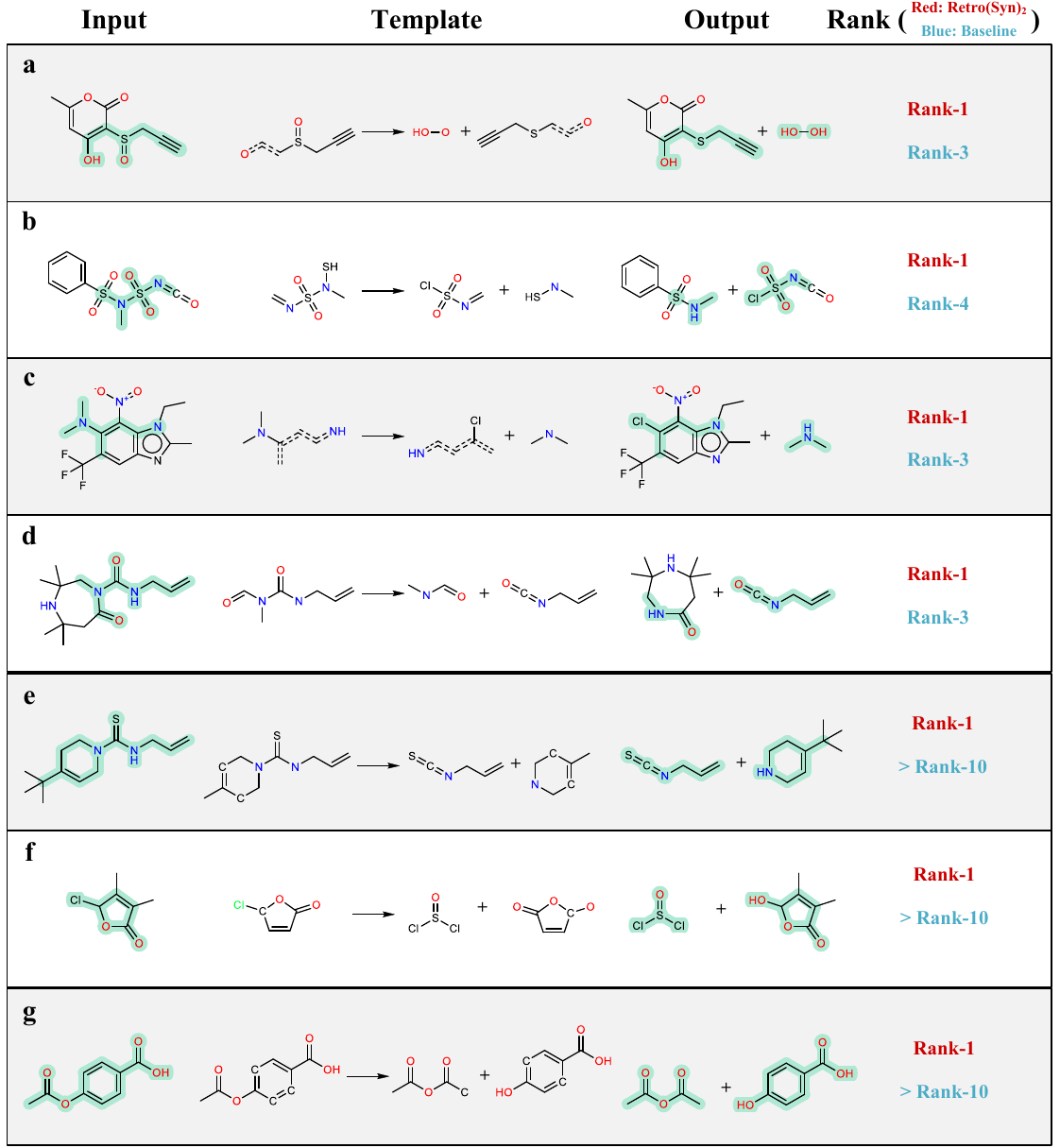}
    \caption{\textbf{Representative examples of Rare2~predictions.} The green part highlights the structure corresponding to the template. \Ours~produces more accurate predictions than Baseline on rare transformations.}\label{fig:examples_rare2}
\end{figure*}

\noindent\textbf{Performance on rare transformations.} Retrosynthesis prediction also faces the challenge of handling rare transformations that involve uncommon reactants, products, or reaction mechanisms, which are underrepresented in the training data. To assess our prediction performance on rare transformations, we create three test subsets from USPTO containing $204,988$, $337,593$, and $438,333$ reactions, where the corresponding template of each reaction appears less than $2$, $5$, and $10$ times, respectively. Correspondingly, these three subsets are denoted as \textbf{Rare-2}, \textbf{Rare-5}, and \textbf{Rare-10}. We conduct an analysis of \Ours~and the baseline both trained on USPTO-50K and report the accuracy on all the test subsets in Fig.\!~\ref{fig:rare_case}. We observe that \Ours~outperforms the baseline across all the subsets, achieving great \textbf{\textit{relative}} improvements. For instance, on the \textbf{Rare-2} subset, \Ours~achieves the relative improvements over the baseline with a top-1 accuracy of 32.2\% and a top-50 accuracy of 23.0\%, respectively. Moreover, we illustrate some representative examples of the Rare-2 subset in Fig.\!~\ref{fig:examples_rare2}. \Ours~produces higher ranking for correct predictions than the baseline. These quantitative and qualitative results indicate that \Ours~better generalizes to rare scenarios.

\begin{figure*}[h]%
    \centering
    \includegraphics[width=\textwidth]{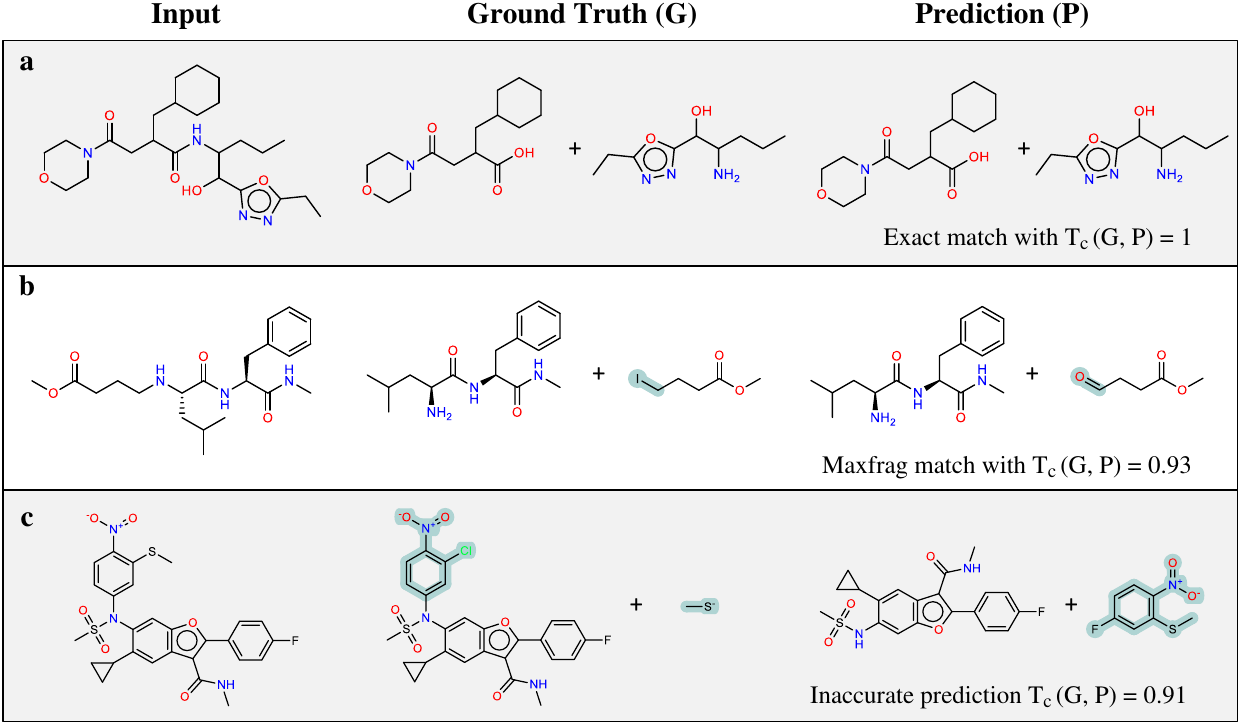}
    \caption{\textbf{Examples of \Ours~predictions.} Representative examples of \textbf{a}, exact match prediction, \textbf{b}, MaxFrag match prediction, and \textbf{c}, inaccurate match prediction are shown. The green part highlights the differences between the ground truth~(G) and the prediction~(P). The molecular similarity is calculated using the ECFP4.}\label{fig:examples}
\end{figure*}

\noindent\textbf{Discussion of prediction results.} The prediction outcomes of \Ours~require a specific comparison for proper evaluation. We  take a much deeper dive into how the predictions are similar to the ground truth by using MaxFrag accuracy and the molecular similarity$_{\!}$~\citep{hendrickson1991concepts,nikolova2003approaches}. Exact match accuracy indicates whether the predicted reactants match the ground truth exactly, while MaxFrag accuracy measures whether the main components of them are identical. Besides, molecular similarity estimates how close the prediction and ground truth are in chemical structure. We show three top-1 predictions of \Ours. Among them, Fig.\!~\ref{fig:examples}a shows an accurate prediction, Fig.\!~\ref{fig:examples}b shows a MaxFrag accurate prediction, where the predicted reactants share the same main fragment as the ground truth (\ie, the minimal part of the reactants to design a retrosynthetic route), and Fig.\!~\ref{fig:examples}c shows an inaccurate prediction. We use the Tanimoto similarity ($T_c$) with ECFP4$_{\!}$~\citep{rogers2010extended} as the molecular fingerprint to quantify the similarity, which ranges from 0 (no overlap) to 1 (complete overlap). Two structures are usually considered similar if $T_c > 0.85$ \citep{maggiora2014molecular}, and we find that even inaccurate predictions from \Ours~usually have high Tanimoto similarity ($T_c=0.91$), indicating that our prediction is might be a feasible outcome from other retrosynthesis routes.

\section{Discussion}\label{sec:discuss}
Retrosynthesis prediction is a challenging task even for experienced chemists due to the huge search space of all possible chemical transformations and the incomplete understanding of the reaction mechanism. Recent machine learning (ML)-based methods have emerged as an efficient tool for chemists in designing synthetic experiments, but their effectiveness heavily hinges on the availability of paired training data (\ie, chemical reactions each consisting of a product-reactant(s) pair), which is expensive to acquire. Furthermore, reaction data is considered a valuable resource by organizations and as a result, its accessibility to the public is severely restricted, creating a major hurdle for researchers. To address these issues, \Ours~utilizes a base model trained on real paired data to generate in-silico reactions from easily accessible unpaired data (\ie, one component of product-reactant(s) pair), thereby facilitating further model training. In this way, the whole framework is self-boosted: pushing the retrosynthesis model to evolve with the in-silico reactions generated by the base model. Besides, ensuring the quality of in-silico reactions is also crucial, which is achieved through a filter process in \Ours.

\Ours~is evaluated on three benchmark datasets and is compared with other state-of-the-art models for single-step retrosynthesis prediction. The experimental results clearly indicate that \Ours~successfully overcomes the training bottleneck caused by the aforementioned issues, \eg, \Ours~achieves a promising 64.9\% top-1 exact match accuracy on USPTO-50K and achieves the top-1 accuracy of 52.3\% in the largest USPTO-Full dataset. Besides, we highlight the superior prediction of \Ours~in almost all reaction classes \eg, \Ours~yields a $9.8\%$ improvement on heterocycle formation class. Moreover, we conduct experiments to show that \Ours~learns more diverse reaction mechanisms, considerably improving the performance on rare transformations. For example, \Ours~achieves the relative improvement of $32.2\%$ over the baseline on top-1 accuracy, indicating that \Ours~has the potential to assist chemists in designing novel routes. In addtion, case studies of prediction show the various possibilities our method can offer for the creation of retrosynthetic routes. 

Despite the promising performance of \Ours, there still remain two challenges in future research: (1) the improvement of \Ours, in large part, relies on the availability and quality of unpaired data, which affects the diversity and chemical plausibility of in-silico reactions. We thus expect that \Ours~could be further enhanced with more sources and methods to collect and preprocess unpaired data. (2) As the number of in-silico reactions grows, it will be more essential to refine the resulting reactions. Therefore, we hypothesize that implementing more efficient and effective filter processes will benefit the advancement of \Ours. With the encouraging experimental results, \Ours~is envisioned to be used as a framework to conquer the training bottleneck of all ML-based methods and stimulate the further development of future ML-based retrosynthesis research.

\section{Methods}\label{sec:method}

\noindent \textbf{Data.} Our models are evaluated on three public benchmark datasets from USPTO curated by$_{\!}$~\cite{lowe2012extraction,Lowe2017}: USPTO-50K$_{\!}$~\citep{schneider2016s}, USPTO-MIT$_{\!}$~\citep{jin2017predicting}, and USPTO-Full$_{\!}$~\citep{dai2019retrosynthesis}. 
\begin{itemize}
    \item USPTO-50K comprises approximately $50,000$ reactions with precise atom mappings between reactants and products. Following$_{\!}$~\cite{liu2017retrosynthetic,dai2019retrosynthesis,zhong2022root}, the 80\%/10\%/10\% of the total 50K reactions are set as train/val/test data. Since the reaction type is usually unknown, we follow$_{\!}$~\cite{zhong2022root} and do not utilize this information for training.
    \item USPTO-MIT (USPTO 480K) dataset contains approximately $400,000$ reactions for training, $30,000$ for validation, and $40,000$ for testing, which is much larger and noisier than the clean USPTO-50K dataset. 
    \item USPTO-FULL is the largest dataset encompassing roughly $1$M chemical reactions, which is built by$_{\!}$~\cite{dai2019retrosynthesis} to verify the scalability of the retrosynthesis model. Following$_{\!}$~\cite{dai2019retrosynthesis,zhong2022root}, Reactions with multiple products are split into individual reactions to ensure that each reaction has only one product, and 1M reactions are divided into train/valid/test sets with sizes of $800$K/$100$K/$100$K respectively.
\end{itemize}

\noindent\textbf{Data representations.} We utilize two molecular representations in this work. 
\begin{itemize}
    \item The Simplified Molecular-Input Line-Entry System (SMILES)$_{\!}$~\citep{weininger1988smiles} is a specification in the form of a line notation for describing the structure of chemical species using short ASCII strings (\eg, c1ccccc1 represents benzene). This representation is widely used as the input and output in most sequence-to-sequence (sequence-based) methods$_{\!}$~\citep{liu2017retrosynthetic,tetko2020state,zhong2022root} for retrosynthesis prediction.
    \item The molecular fingerprint is a bit-vector encoding the physicochemical or structural properties of the molecule, which is usually used for synthesis design$_{\!}$~\citep{segler2017neural}, similarity searching$_{\!}$~\citep{willett1998chemical}, and virtual screening$_{\!}$~\citep{cereto2015molecular,muegge2016overview}, \etc. The most used ones are Extended-Connectivity Fingerprint (ECFP)$_{\!}$~\citep{rogers2010extended} and Maccs-Keys$_{\!}$~\citep{durant2002reoptimization}. The molecular fingerprint is utilized in this work to quantify the molecular similarity between two molecules, indicating their closeness.
\end{itemize}

\noindent \textbf{Problem formulation.} The single-step retrosynthesis prediction task aims to predict precursors by inputting a molecule of interest. ML-based methods rely on the dataset of paired data in the product and corresponding reactant(s), denoted as $\mathit{R}\!=\!\{(x_{n},y_{n})\}_{n}$, where $x_{n}$ represents the product and $y_{n}$ denotes the corresponding reactant(s). In this work, both reactants and products are represented by SMILES. Given a product sequence $x$, a sequence-based method learns a model $f_{x\myarrow[0.3]y}$ to obtain the corresponding reactant(s) sequence $y$.

\noindent\textbf{Baseline.} The baseline of our \textbf{Retro}Synthesis \textbf{W}ith \textbf{I}n \textbf{S}ilico r\textbf{E}actions (\Ours) framework only uses real paired data for training, adopting the vanilla transformer$_{\!}$~\citep{vaswani2017attention} as the network architecture, and using Root-aligned SMILES (R-SMILES)$_{\!}$~\citep{zhong2022root} as the SMILES augmentation strategy. 
Transformer consists of an encoder-decoder architecture where the encoder maps the input sequence to a latent space, and the decoder decodes the output sequence from the latent space in an autoregressive manner. 
R-SMILES is a tightly aligned one-to-one mapping between the product and the reactant(s) sequence for more efficient retrosynthesis prediction. It adopts the same atom as the root (the starting atom) to transform molecules into SMILES sequences for the product and the corresponding reactant(s).

\noindent\textbf{In-silico reaction generation.} First, \Ours~uses real paired data (\eg, reactions in USPTO-50K) to train a base forward synthesis model and a base retrosynthesis model. Then, \Ours~collects unpaired data, the amount of which typically far exceeds the amount of paired data, from one of two sources: one containing unpaired reactants and one containing unpaired products. The reactants are derived from the USPTO 2001-2016 applications containing $1,939,254$ raw reactions. Although these data are paired, the reactant(s) component of each reaction is only utilized to verify the effectiveness of our proposed framework.
We preprocess raw reactions by removing duplicates, reactions with incorrect atom mappings, and reactions with multiple products (which we split into separate ones). Reactions that appear in the validation or test set of the existing dataset are also excluded. Then, the base forward synthesis model is used to produce in-silico products from the unpaired reactants, leading to more in-silico reactions. Also, the unpaired products are obtained by randomly sampling $4$M molecules from PubChem or $20$M from ZINC, respectively. The base retrosynthesis model is used to generate in-silico reactions following a similar procedure as before. The base model performs beam-search decoding for the newly introduced unpaired data and select the best one as the in-silico data. Moreover, a filter process is adopted to enhance the quality of the in-silico reactions, which contains a template matching step and a molecular similarity comparison step. For the first step, we use RDChiral$_{\!}$~\citep{coley2019rdchiral} to extract $1,808,176$ templates from USPTO. We select those that appear more than $5$ times (\ie, $43,710$ unique templates) to do template matching. For the second step, we set the threshold to be $0.55$ to do a molecular similarity comparison. Ultimately, \Ours~uses roughly $89\%$ of in-silico reactions through the two filtering steps to achieve a better performance.

\noindent\textbf{Training details.} As in$_{\!}$~\cite{tetko2020state,seo2021gta,zhong2022root}, we apply the SMILES augmentation during training for our \Ours~framework: 20\,\!$\times$ SMILES augmentation for USPTO-50K$_{\!}$~\citep{schneider2016s}, 5\,\!$\times$ SMILES augmentation for USPTO-MIT$_{\!}$~\citep{jin2017predicting}, and 5\,\!$\times$ SMILES augmentation for USPTO-Full$_{\!}$~\citep{dai2019retrosynthesis}. We use the OpenNMT framework$_{\!}$~\citep{klein2017opennmt} and PyTorch$_{\!}$~\citep{paszke2019pytorch} to build the transformer model. Following$_{\!}$~\cite{irwin2022chemformer, zhong2022root}, we use the masking strategy to pretrain the model before training. During training, we employ the Adam optimizer$_{\!}$~\citep{kingma2014adam} with $\beta_1=0.9$, $\beta_2=0.998$  for loss minimization and apply dropout$_{\!}$~\citep{srivastava2014dropout} to the whole model at a rate of $0.1$. The starting learning rate is set to $1.0$ and noam$_{\!}$~\citep{vaswani2017attention} is used as the learning rate decay scheme.

\noindent\textbf{Evaluation procedure.} We use the top-\textit{k} exact match accuracy as the evaluation metric to assess the performance of each model, where the \textit{k} ranges from \{1, 3, 5, 10, 20, 50\}. This metric is widely used in existing studies$_{\!}$~\citep{liu2017retrosynthetic,kim2021valid,karpov2019transformer,sacha2021molecule,wang2021retroprime}, which measures the ratio that one of the top-\textit{k} predicted results exactly match the ground truth. We additionally adopt the top-\textit{k} MaxFrag accuracy introduced by$_{\!}$~\cite{tetko2020state} for retrosynthesis. Compared with the exact match accuracy, the MaxFrag accuracy focuses on main compound transformations, which are the minimal information required to get a retrosynthesis route. As in$_{\!}$~\cite{tetko2020state,seo2021gta,zhong2022root}, we apply the same SMILES augmentations at the evaluation stage as during training.

\backmatter

\section{Supplementary information} 
\begin{appendices}
    \makeatletter
    \renewcommand \thesection{S\@arabic\c@section}
    \renewcommand\thetable{S\@arabic\c@table}
    \renewcommand \thefigure{S\@arabic\c@figure}
    \makeatother
\noindent\textbf{Details of baseline.} In this work, we adopt the vanilla transformer \citep{vaswani2017attention} as the network architecture. A typical transformer model consists of two major parts called encoder and decoder. There are several identical layers of transformer encoder and each has three separate blocks, named as ``Layer Norm", ``Multi-head Self Attention (MSA)", and ``Feedforward Network (FFN)". Among them, the attention mechanism is the most critical part of transformer, where three different vectors Keys($K$), Queries($Q$) and Values($V$) of dimension $d$ are employed for each input token. For computing the self attention metric, the dot product of Queries and all the Keys are calculated and scaled by $1/\sqrt{d}$ in order to prevent the dot products from generating very large numbers. This matrix is then converted into a probability matrix through the $softmax$ function and is multiplied to the Values to produce the attention metric as follows:

\begin{equation}
    Attention = softmax(\frac{QK^T}{\sqrt{d}})V.
\end{equation}
Besides, the baseline utilizes Root-aligned SMILES (R-SMILES) \citep{zhong2022root} as our SMILES augmentation strategy. R-SMILES has a better augmentation effect because it adopts a tightly aligned one-to-one mapping between the product and the reactant to predict retrosynthesis more effectively. Specifically, it adopts the same atom as the root (\ie, the starting atom) of the SMILES strings for both the products and the reactants, which successfully resolves the one-to-many problem in random augmentation and enriches the SMILES representation compared to using canonical SMILES. 

\begin{figure}
    \centering
    \includegraphics[width=\textwidth]{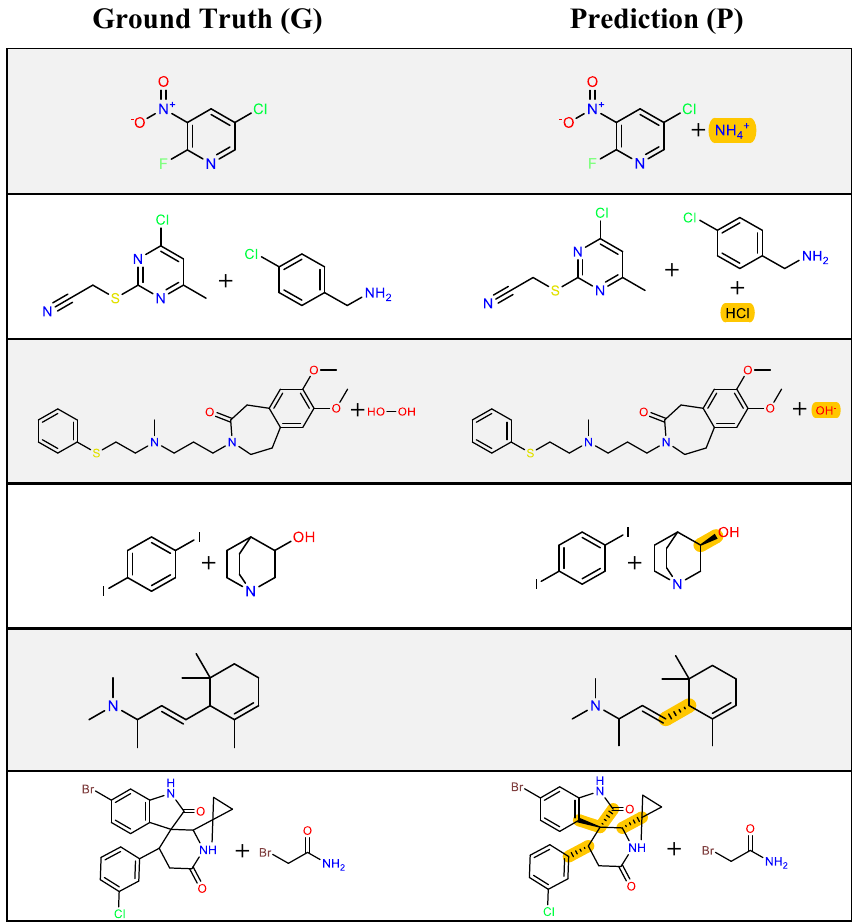}
    \caption{Predictions with very high molecule similarity ($Tc\geq0.95$) but inconsistent with the ground truth. We highlight differences between the ground truth and the prediction.}
    \label{fig:example-more}
\end{figure}

\begin{table}[t]
    \caption{Iterative training yields high-quality in-silico reactions and accurate prediction.} 
    \label{tab:iterative}
    \setlength\tabcolsep{13pt}
      \resizebox{\linewidth}{!}{
    \begin{tabular}{@{}lcccccc}
    \toprule
    Method & \textit{k} = 1 & 3    & 5    & 10   & 20   & 50   \\ \midrule
    Baseline    &63.8	&83.0	&87.6	&91.7	&94.1	&95.1 \\
    \Ours~+Iterative     &\textbf{64.9}	&\textbf{83.8}	&\textbf{88.0}	&\textbf{91.9}	&\textbf{94.3}	& \textbf{96.1} \\

    \bottomrule
    \end{tabular} }
\end{table}

\noindent\textbf{Iterative training.} \Ours~is a self-boosting framework and could benefit from iterative training. Specifically, a better base model will result in better in-silico reactions, leading to improved predictions for retrosynthesis. If we can build a better base model with the in-silico reactions, then we can continue repeating this process: utilizing the base model to generate in-silico reactions, and building an even better base model with these reactions to generate higher-quality reactions for training. In other words, the key idea is to build a better base model with previous in-silico reactions for iteratively augmenting real paired data. Table~\ref{tab:iterative} suggests that adding one more iteration enhances the prediction performance of the retrosynthesis model (\eg, +1.1\% top-1 accuracy). However, iterative training also has several drawbacks, such as significantly increasing the training and generation time with too many iterations, and introducing more biases during iterative training. 

\noindent\textbf{Discussions of highly scored inaccurate predictions.} Two chemical structures are typically considered similar if the Tanimoto coefficient ($T_c$) is above 0.85$_{\!}$~\citep{maggiora2014molecular}. We previously presented an inaccurate prediction with a high similarity ($T_c = 0.91$) in the main manuscript for a more proper evaluation, which demonstrates the prediction diversity of \Ours. As illustrated in Fig.\!~\ref{fig:example-more}, we provide more examples from the USPTO-50K test set with higher similarity ($T_c \geq 0.95$), highlighting some challenges faced by machine learning (ML)-based methods: (1) the tendency of ML-based models to generate unnecessary reagents like {NH$_{4}^+$}, {HCL}, and {OH$^-$} due to learning bias; (2) the failure of models to accurately represent molecular information of stereochemistry, such as using incorrect symbols (/ or $\backslash$) to denote directional single bonds adjacent to a double bond or creating accurate molecules but with incorrect chirality (\eg, C@H v.s. C@@H).

\begin{figure}[htbp]
    \centering
    \subfigure{\includegraphics[width=0.49\linewidth]{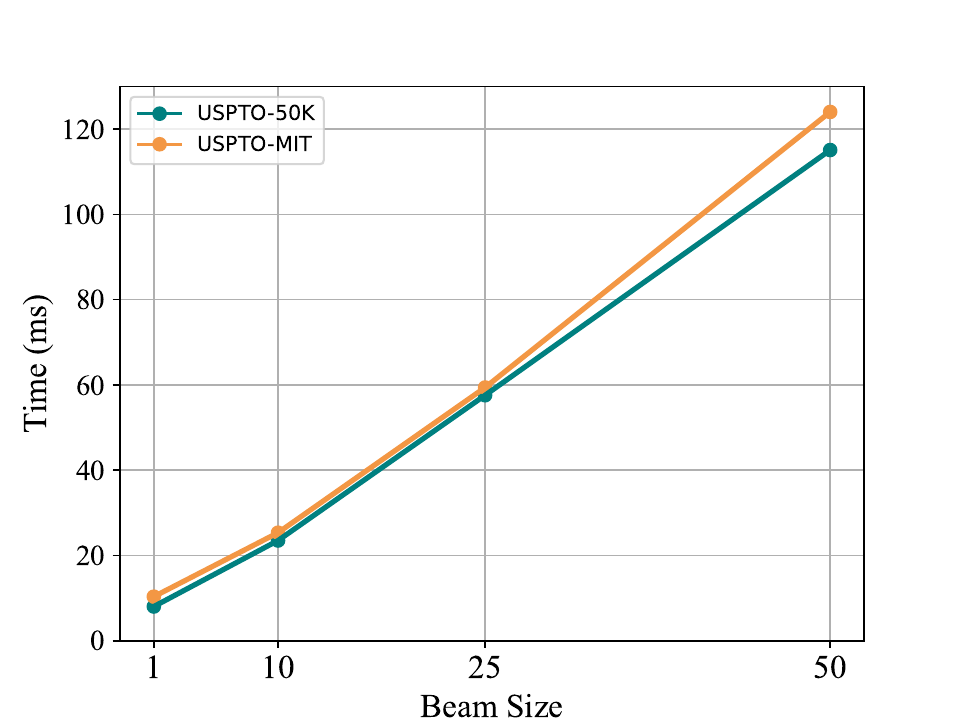}\label{fig:left2}}
    \put(-149, 127){\fontsize{6.5pt}{1em}\selectfont \textbf{a}}
    \hfill
    \subfigure{\includegraphics[width=0.49\linewidth]{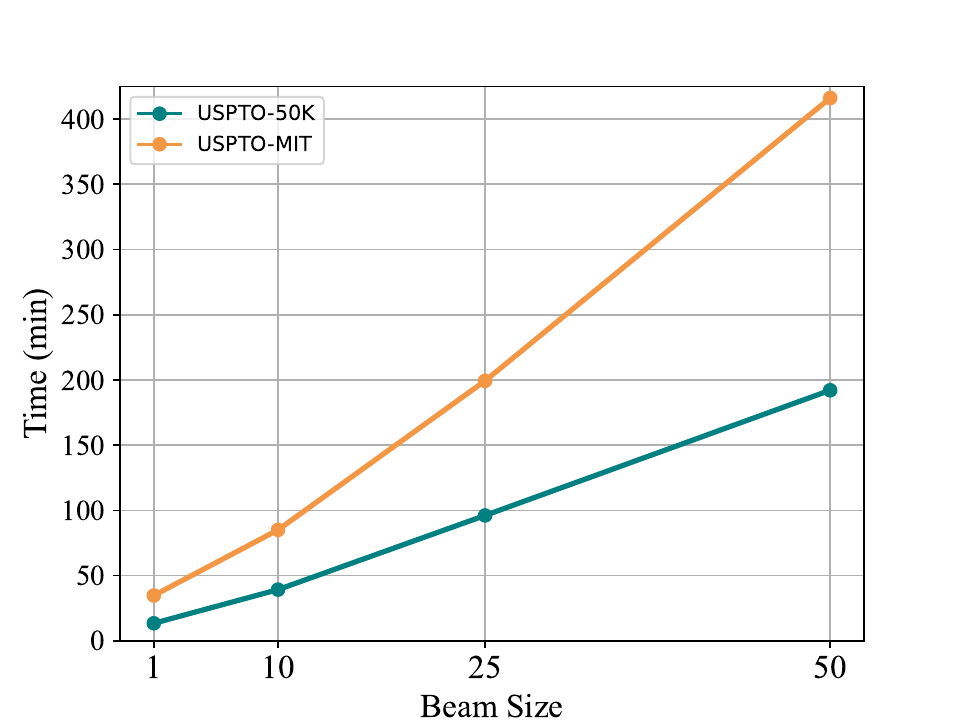}\label{fig:right2}}
    \put(-149, 127){\fontsize{6.5pt}{1em}\selectfont \textbf{b}}
    \caption{\textbf{Inference time of \Ours~with different beam sizes.} The inference time is measured with \textbf{a}, ``Time per product'' and \textbf{b}, ``Total time''. ``Time per product'' measures the time required to generate one reactant(s) from a given product using \Ours. ``Total time'' is the time to decode the whole test set into prediction results.}\label{fig:infer}
\end{figure}

\noindent\textbf{Computational and memory efficiency.} The proposed \Ours~framework prioritizes memory and computational efficiency during inference, which enables further applications, such as multistep retrosynthesis planning. \Ours~employs a transformer architecture with approximately $44.5$M parameters as the sequence-based model for USPTO-50K and USPTO-MIT. Compared with previous transformer-based method such as RetroPrime$_{\!}$~\citep{wang2021retroprime} having $75.4$M parameters, \Ours~is more lightweight and easy to deploy. Fig.\!~\ref{fig:infer} illustrates the inference speed on different datasets, measured with a single GPU (GeForce RTX 4090). The time per product varies with different beam sizes. On USPTO-50K, it is between $8.03$ ms and $115.07$ ms, while on USPTO-MIT, which has longer sequences, it is between $10.35$ ms and $123.98$ ms. The total time also depends on the beam size and the dataset. For the USPTO-50K test set with $10,014$ products, it varies from $13.41$ min to $192.06$ min. For the USPTO-MIT test set with $201,325$ products, the range is from $34.73$ min to $416.11$ min. These experimental results highlight the computational and memory efficiency of \Ours.
\end{appendices}

\bibliography{sn-bibliography}
%\bibliography{bibs}


\end{document}